\documentclass[conference]{IEEEtran}
\IEEEoverridecommandlockouts
\usepackage{cite}
\usepackage{amsmath,amssymb,amsfonts}
\usepackage{algorithmic}
\usepackage{graphicx}
\usepackage{textcomp}
\usepackage[usenames,dvipsnames]{xcolor}
\usepackage{balance}
\usepackage[hidelinks,
            colorlinks,
            linkcolor={red!80!black},
            citecolor={green!80!black},
            urlcolor={blue!80!black}]{hyperref}
\usepackage{booktabs,multirow,tabularx,threeparttable}
\usepackage{courier}
\usepackage[T1]{fontenc}
\def\BibTeX{{\rm B\kern-.05em{\sc i\kern-.025em b}\kern-.08em
    T\kern-.1667em\lower.7ex\hbox{E}\kern-.125emX}}
\begin{document}

\title{Fine-grained Sentiment Classification using BERT}

\makeatletter
\newcommand{\linebreakand}{
  \end{@IEEEauthorhalign}
  \hfill\mbox{}\par
  \mbox{}\hfill\begin{@IEEEauthorhalign}
}
\makeatother
\author{
    \IEEEauthorblockN{
        Manish Munikar\IEEEauthorrefmark{1},
        Sushil Shakya\IEEEauthorrefmark{2} and
        Aakash Shrestha\IEEEauthorrefmark{3}
    }
    \IEEEauthorblockA{
        {\it Department of Electronics and Computer Engineering}\\
        {\it Pulchowk Campus, Institute of Engineering, Tribhuvan University}\\
        Lalitpur, Nepal\\
        \IEEEauthorrefmark{1}{\tt 070bct520@ioe.edu.np}, 
        \IEEEauthorrefmark{2}{\tt 070bct547@ioe.edu.np}, 
        \IEEEauthorrefmark{3}{\tt 070bct501@ioe.edu.np}
    }
}

\maketitle

\begin{abstract}
Sentiment classification is an important process in understanding people's perception towards a product, service, or topic. Many natural language processing models have been proposed to solve the sentiment classification problem. However, most of them have focused on binary sentiment classification. In this paper, we use a promising deep learning model called BERT to solve the fine-grained sentiment classification task. Experiments show that our model outperforms other popular models for this task without sophisticated architecture. We also demonstrate the effectiveness of transfer learning in natural language processing in the process.
\end{abstract}

\section{Introduction}

Sentiment classification is a form of text classification in which a piece of text has to be classified into one of the predefined sentiment classes. It is a supervised machine learning problem. In binary sentiment classification, the possible classes are positive and negative. In fine-grained sentiment classification, there are five classes (very negative, negative, neutral, positive, and very positive). Fig \ref{fig:sentiment} shows a black-box view of a fine-grained sentiment classifier model.

\begin{figure}[ht]
    \centering
    \includegraphics[width=0.9\linewidth]{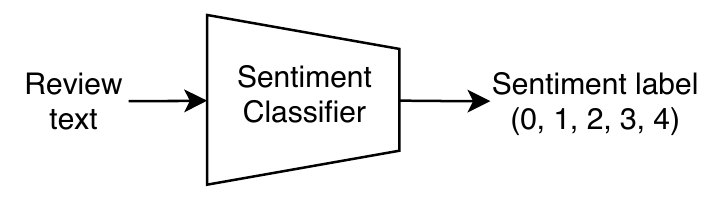}
    \caption{High-level black-box view of a sentiment classifier showing its input and output.}
    \label{fig:sentiment}
\end{figure}

Sentiment classification model, like any other machine learning model, requires its input to be a fixed-sized vector of numbers. Therefore, we need to convert a text---sequence of words represented as ASCII or Unicode---into a fixed-sized vector that encodes the meaningful information of the text. Many statistical and deep learning NLP models have been proposed just for that. Recently, there has been an explosion of developments in NLP as well as other deep learning architectures.

While transfer learning (pretraining and finetuning) has become the de-facto standard in computer vision, NLP is yet to utilize this concept fully. However, neural language models such as word vectors\cite{wordvec}, paragraph vectors\cite{doc2vec}, and GloVe\cite{glove} have started the transfer learning revolution in NLP. Recently, Google researchers published BERT (Bidirectional Encoder Representations from Transformers)\cite{bert}, a deep bidirectional language model based on the Transformer architecture\cite{attention}, and advanced the state-of-the-art in many popular NLP tasks. In this paper, we use the pretrained BERT model and fine-tune it for the fine-grained sentiment classification task on the Stanford Sentiment Treebank (SST) dataset.

The rest of the paper is organized into six sections. In Section~\ref{sec:motivation}, we mention our motivation for this work. In Section~\ref{sec:related}, we discuss related works. In Section~\ref{sec:data}, we describe the dataset we performed our experiments on. We explain our model architecture and methodology in detail in Section~\ref{sec:method}. Then we present and analyze our results in Section~\ref{sec:result}. Finally, we provide our concluding remarks in Section~\ref{sec:conclusion}.
\section{Motivation}\label{sec:motivation}

We have been working on replicating the different research paper results for sentiment analysis, especially on the fine-grained Stanford Sentiment Treebank (SST) dataset. After the popularity of BERT, researchers have tried to use it on different NLP tasks, including binary sentiment classification on SST-2 (binary) dataset, and they were able to obtain state-of-the-art results as well. But we haven't yet found any experimentation done using BERT on the SST-5 (fine-grained) dataset. Because BERT is so powerful, fast, and easy to use for downstream tasks, it is likely to give promising results in SST-5 dataset as well. This became the main motivation for pursuing this work.
\section{Related Work}\label{sec:related}

Sentiment classification is one of the most popular tasks in NLP, and so there has been a lot of research and progress in solving this task accurately. Most of the approaches have focused on binary sentiment classification, most probably because there are large public datasets for it such as the IMDb movie review dataset\cite{imdb-dataset}. In this section, we only discuss some significant deep learning NLP approaches applied to sentiment classification.

The first step in sentiment classification of a text is the embedding, where a text is converted into a fixed-size vector. Since the number of words in the vocabulary after tokenization and stemming is limited, researchers first tackled the problem of learning word embeddings. The first promising language model was proposed by Mikolov et al.\cite{wordvec}. They trained continuous semantic representation of words from large unlabeled text that could be fine-tuned for downstream tasks. Pennington et al.\cite{glove} used a co-occurrence matrix and only trained on non-zero elements to efficiently learn semantic word embeddings. Bojanowski et al.\cite{fasttext} broke words into character $n$-grams for smaller vocabulary size and fast training.

The next step is to combine a variable number of word vectors into a single fixed-size document vector. The trivial way is to take the sum or the average, but they don't lose the ordering information of words and thus don't give good results. Tai et al.\cite{rnn-sentiment} used recursive neural networks to compute vector representation of sentences by utilizing the intrinsic tree structure of natural language sentences. Socher et al.\cite{rntn} introduced a tensor-based compositionaity function for better interaction between child nodes in recursive networks. They also introduced the Stanford Sentiment Treebank (SST) dataset for fine-grained sentiment classification. Tai et al.\cite{lstm} applied various forms of long short-term memory (LSTM) networks and Kim\cite{cnn} applied convolutional neural networks (CNN) towards sentiment classification.

All of the approaches mentioned above are context-free, i.e., they generate single word embedding for each word in the vocabulary. For instance, ``bank`` would have the same representation in ``bank deposit`` and ``river bank``. Recent language model research has been trying to train contextual embeddings. Peters et al.\cite{elmo} extracted context-sensitive features from left-to-right and right-to-left LSTM-based language model. Devlin et al.\cite{bert} proposed BERT (Bidirectional Encoder Representations from Transformers), an attention-based Transformer architecture\cite{attention}, to train deep bidirectional representations from unlabeled texts. Their architecture not only obtains state-of-the-art results on many NLP tasks but also allows a high degree of parallelism since it is not based on sequential or recurrent connections.

\section{Dataset}\label{sec:data}

Stanford Sentiment Treebank (SST)\cite{rntn} is one of the most popular publicly available datasets for fine-grained sentiment classification task. It contains 11,855 one-sentence movie reviews extracted from Rotten Tomatoes. Not only that, each sentence is also parsed by the Stanford constituency parser\cite{stanford-parser} into a tree structure with the whole sentence as the root node and the individual words as leaf nodes. Moreover, each node is labeled by at least three humans. In total, SST contains 215,154 unique manually labeled texts of varying lengths. Fig \ref{fig:sst-sample} shows a sample review from the SST dataset in a parse-tree structure with all its nodes labeled. Therefore, this dataset can be used to train models to learn the sentiment of words, phrases, and sentences together.

\begin{figure}[ht]
    \centering
    \fbox{\includegraphics[width=0.9\linewidth]{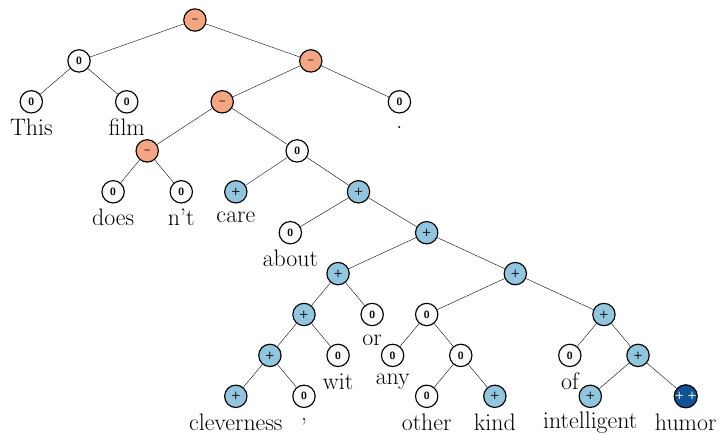}}
    \caption{A sample sentence from the SST dataset. (Source: Adapted from  \cite{rntn}.)}
    \label{fig:sst-sample}
\end{figure}

There are five sentiment labels in SST: 0 (very negative), 1 (negative), 2 (neutral), 3 (positive), and 4 (very positive). If we only consider positivity and negativity, we get the binary SST-2 dataset. If we consider all five labels, we get SST-5. For this research, we evaluate the performance of various models on all nodes as well as on just the root nodes, and on both SST-2 and SST-5.
\section{Methodology}\label{sec:method}

Sentiment classification takes a natural language text as input and outputs a sentiment score $\in \left\{0,1,2,3,4\right\}$. Our method has three stages from input sentence to output score, which are described below. We use pretrained BERT model to build a sentiment classifier. Therefore, in this section, we briefly explain BERT and then describe our model architecture.

\subsection{BERT}

BERT (Bidirectional Encoder Representations from Transformers is an embedding layer designed to train deep bidirectional representations from unlabeled texts by jointly conditioning on both left and right context in all layers. It is pretrained from a large unsupervised text corpus (such as Wikipedia dump or BookCorpus) using the following objectives:

\begin{itemize}
    \item \emph{Masked word prediction:}
        In this task, 15\% of the words in the input sequence are masked out, the entire sequence is fed to a deep bidirectional Transfomer\cite{attention} encoder, and then the model learns to predict the masked words.
    \item \emph{Next sentence prediction:}
        To learn the relationship between sentences, BERT takes two sentences $A$ and $B$ as inputs and learns to classify whether $B$ actually follows $A$ or is it just a random sentence.
\end{itemize}

Unlike traditional sequential or recurrent models, the attention architecture processes the whole input sequence at once, enabling all input tokens to be processed in parallel. The layers of BERT architecture are visualized in Fig~\ref{fig:bert}. Pretrained BERT model can be fine-tuned with just one additional layer to obtain state-of-the-art results in a wide range of NLP tasks\cite{bert}.

\begin{figure}[ht]
    \centering
    \includegraphics[width=0.7\linewidth]{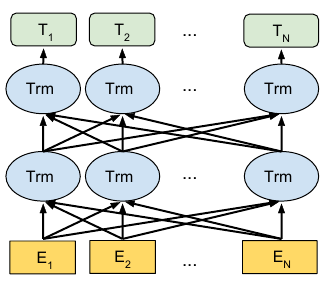}
    \caption{BERT Architecture, where $E_n$ is the $n$-th token in the input sequence, Trm is the Transformer block, and $T_n$ is the corresponding output embedding. (Source: Adapted from \cite{bert}.)}
    \label{fig:bert}
\end{figure}

There are two variants for BERT models: BERT\textsubscript{BASE} and BERT\textsubscript{LARGE}. The difference between them is listed in Table~\ref{tab:base-large}.

\begin{table}[ht]
    \renewcommand{\arraystretch}{1.2}
    \caption{BERT\textsubscript{BASE} vs. BERT\textsubscript{LARGE}.}
    \label{tab:base-large}
    \centering
    \begin{tabular}{lcc}
        \toprule
         & BERT\textsubscript{BASE} & BERT\textsubscript{LARGE} \\
         \midrule
        No. of layers (Transformer blocks) & 12 & 24 \\
        No. of hidden units & 768 & 1024 \\
        No. of self-attention heads & 12 & 16 \\
        Total trainable parameters & 110M & 340M \\
        \bottomrule
    \end{tabular}
\end{table}

\subsubsection{Input format}

BERT requires its input token sequence to have a certain format. First token of every sequence should be {\tt [CLS]} (classification token) and there should be a {\tt [SEP]} token (separation token) after every sentence. The output embedding corresponding to the {\tt [CLS]} token is the sequence embedding that can be used for classifying the whole sequence.

\subsection{Preprocessing}

We perform the following preprocessing steps on the review text before we feed them into out model.

\subsubsection{Canonicalization} First, we remove all the digits, punctuation symbols and accent marks, and convert everything to lowercase.

\subsubsection{Tokenization} We then tokenize the text using the WordPiece tokenizer\cite{wordpiece}. It breaks the words down to their prefix, root, and suffix to handle unseen words better. For example, {\tt playing} $\rightarrow$ {\tt play} + {\tt \#\#ing}.

\subsubsection{Special token addition} Finally, we add the {\tt [CLS]} and {\tt [SEP]} tokens at the appropriate positions.

\subsection{Proposed Architecture}

We build a simple architecture with just a dropout regularization\cite{dropout} and a softmax classifier layers on top of pretrained BERT layer to demonstrate that BERT can produce great results even without any sophisticated task-specific architecture.

\begin{figure}[ht]
    \centering
    \includegraphics[width=0.4\linewidth]{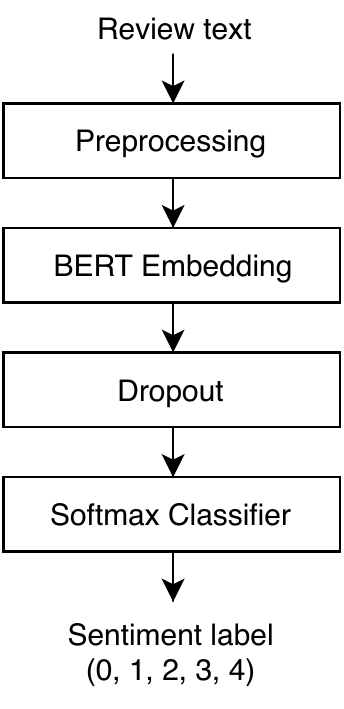}
    \caption{Proposed architecture for fine-grained sentiment classification.}
    \label{fig:architecture}
\end{figure}

Fig~\ref{fig:architecture} shows the overall architecture of our model. There are four main stages. The first is the proprocessing step as described earlier. Then we compute the sequence embedding from BERT. We then apply dropout with a probability factor of $0.1$ to regularize and prevent overfitting. Dropout is only applied in training phase and not in inference phase. Finally, the softmax classification layer will output the probabilities of the input text belonging to each of the class labels such that the sum of the probabilities is $1$. The softmax layer is just a fully connected neural network layer with the softmax activation function. The softmax function $\sigma : \mathbb{R}^K \rightarrow \mathbb{R}^K$ is given in \eqref{eq:softmax}.
\begin{equation}
    \sigma(\mathbf{z})_i = \frac{e^{z_i}}{\sum_{j=1}^K e^{z_j}}
    \text{ for } i=1,\ldots,K
    \label{eq:softmax}
\end{equation}
where $\mathbf{z} = (z_1,\ldots,z_K) \in \mathbb{R}^K$ is the intermediate output of the softmax layer (also called logits). The output node with the highest probability is then chosen as the predicted label for the input.
\section{Experiments and Results}\label{sec:result}

In this section, we discuss the results of our model and compare with it some of the popular models that solve the same problem, i.e., sentiment classification on the SST dataset.

\subsection{Comparison Models}



\subsubsection{Word embeddings}
In this method, the word vectors pretrained on large text corpus such as Wikipedia dump are averaged to get the document vector, which is then fed to the sentiment classifier to compute the sentiment score.



\subsubsection{Recursive networks}
Various types of recursive neural networks (RNN) have been applied on SST\cite{rntn}. We compare our results with the standard RNN and the more sophisticated RNTN. Both of them were trained on SST from scratch, without pretraining.

\subsubsection{Recurrent networks}
Sophisticated recurrent networks such as left-to-right and bidrectional LSTM networks have also been applied on SST\cite{lstm}.

\subsubsection{Convolutional networks}
In this approach, the input sequences were passed through a 1-dimensional convolutional neural network as feature extractors\cite{cnn}.

\subsection{Evaluation Metric}

Since the dataset has roughly balanced number of samples of all classes, we directly use the accuracy measure to evaluate the performance of our model and compare it with other models. The accuracy is defined simply as follows:
\begin{equation}
    \text{accuracy} =
    \frac{
        \text{number of correct predictions}
    }{
        \text{total number of predictions}
    } \in [0,1]
\end{equation}

\subsection{Results}

The result and comparisons are shown in Table \ref{tab:result}. It shows the accuracy of various models on SST-2 and SST-5. It includes results for all phrases as well as for just the root (whole review). We can see that our model, despite being a simple architecture, performs better in terms of accuracy than many popular and sophisticated NLP models.

\begin{table}
    \renewcommand{\arraystretch}{1.2}
    \begin{center}
    \begin{threeparttable}
    \caption{Accuracy (\%) of our models on SST dataset compared to other models.\tnote{1}}
    \label{tab:result}
    \begin{tabular}{p{4cm}cccc}
        \toprule
        \multirow{2}{*}{Model} & \multicolumn{2}{c}{SST-2} & \multicolumn{2}{c}{SST-5}\\
        \cmidrule(lr){2-3}
        \cmidrule(lr){4-5}
        & All & Root & All & Root\\
        \midrule
        Avg word vectors\cite{rntn} & 85.1 & 80.1 & 73.3 & 32.7 \\
        RNN\cite{rnn-sentiment} & 86.1 & 82.4 & 79.0 & 43.2 \\
        RNTN\cite{rntn} & 87.6 & 85.4 & 80.7 & 45.7 \\
        Paragraph vectors\cite{doc2vec} & -- & 87.8 & -- & 48.7\\
        LSTM\cite{lstm} & -- & 84.9 & -- & 46.4\\
        BiLSTM\cite{lstm} & -- & 87.5 & -- & 49.1\\
        CNN\cite{cnn} & -- & 87.2 & -- & 48.0\\
        \midrule
        BERT\textsubscript{BASE} & 94.0 & 91.2 & 83.9 & 53.2 \\
        BERT\textsubscript{LARGE} & {\bf 94.7} & {\bf 93.1} & {\bf 84.2} & {\bf 55.5} \\
        \bottomrule       
    \end{tabular}
    \begin{tablenotes}
        \item[1] Some values are blank in ``All'' columns because the original authors of those paper did not publish their result on all phrases.
    \end{tablenotes}
    \end{threeparttable}
    \end{center}
\end{table}
\section{Conclusion}\label{sec:conclusion}

In this paper, we used the pretrained BERT model and fine-tuned it for the fine-grained sentiment classification task on the SST dataset. Even with such a simple downstream architecture, our model was able to outperform complicated architectures like recursive, recurrent, and convolutional neural networks. Thus, we have demonstrated the transfer learning capability in NLP enabled by deep contextual language models like BERT.

\section*{Acknowledgment}

We would like to express our gratitude towards Prof. Dr. Shashidhar Ram Joshi for his invaluable advice and guidance on this paper. We also thank all the helpers and reviewers for their valuable input to this work.
\balance
\bibliographystyle{IEEEtran}
\bibliography{main}

\end{document}